\documentclass[10pt,twocolumn,letterpaper]{article}

\usepackage[pagenumbers]{cvpr} 

\usepackage{amssymb}
\usepackage{pifont}
\newcommand{\cmark}{\ding{51}}%
\newcommand{\xmark}{\ding{55}}%
\usepackage{tablefootnote}
\usepackage{multirow}

\definecolor{cvprblue}{rgb}{0.21,0.49,0.74}
\usepackage[pagebackref,breaklinks,colorlinks,allcolors=cvprblue]{hyperref}

\def\paperID{16801} 
\def\confName{CVPR}
\def\confYear{2025}

\title{Every Image Listens, Every Image Dances: Music-Driven Image Animation}

\author{
    Zhikang Dong$^{1}$\thanks{Work done during the internship at Bytedance.} \quad
    Weituo Hao$^2$\quad
    Ju-Chiang Wang$^2$\quad
    Peng Zhang$^{3}$\thanks{Work done at Bytedance.}\quad
    Pawel Polak$^1$\quad
    \\
    \\
    $^1$Stony Brook University \quad
    $^2$Bytedance \quad
    $^3$Apple
}

\begin{document}
\maketitle
\begin{abstract}
Image animation has become a promising area in multimodal research, with a focus on generating videos from reference images. While prior work has largely emphasized generic video generation guided by text, music-driven dance video generation remains underexplored. In this paper, we introduce MuseDance, an innovative end-to-end model that animates reference images using both music and text inputs. This dual input enables MuseDance to generate personalized videos that follow text descriptions and synchronize character movements with the music. Unlike existing approaches, MuseDance eliminates the need for complex motion guidance inputs, such as pose or depth sequences, making flexible and creative video generation accessible to users of all expertise levels. To advance research in this field, we present a new multimodal dataset comprising 2,904 dance videos with corresponding background music and text descriptions. Our approach leverages diffusion-based methods to achieve robust generalization, precise control, and temporal consistency, setting a new baseline for the music-driven image animation task.
\end{abstract}    
\section{Introduction}
\label{sec:intro}

\begin{figure*}[!th]
  \centering
  \includegraphics[width=1\textwidth]{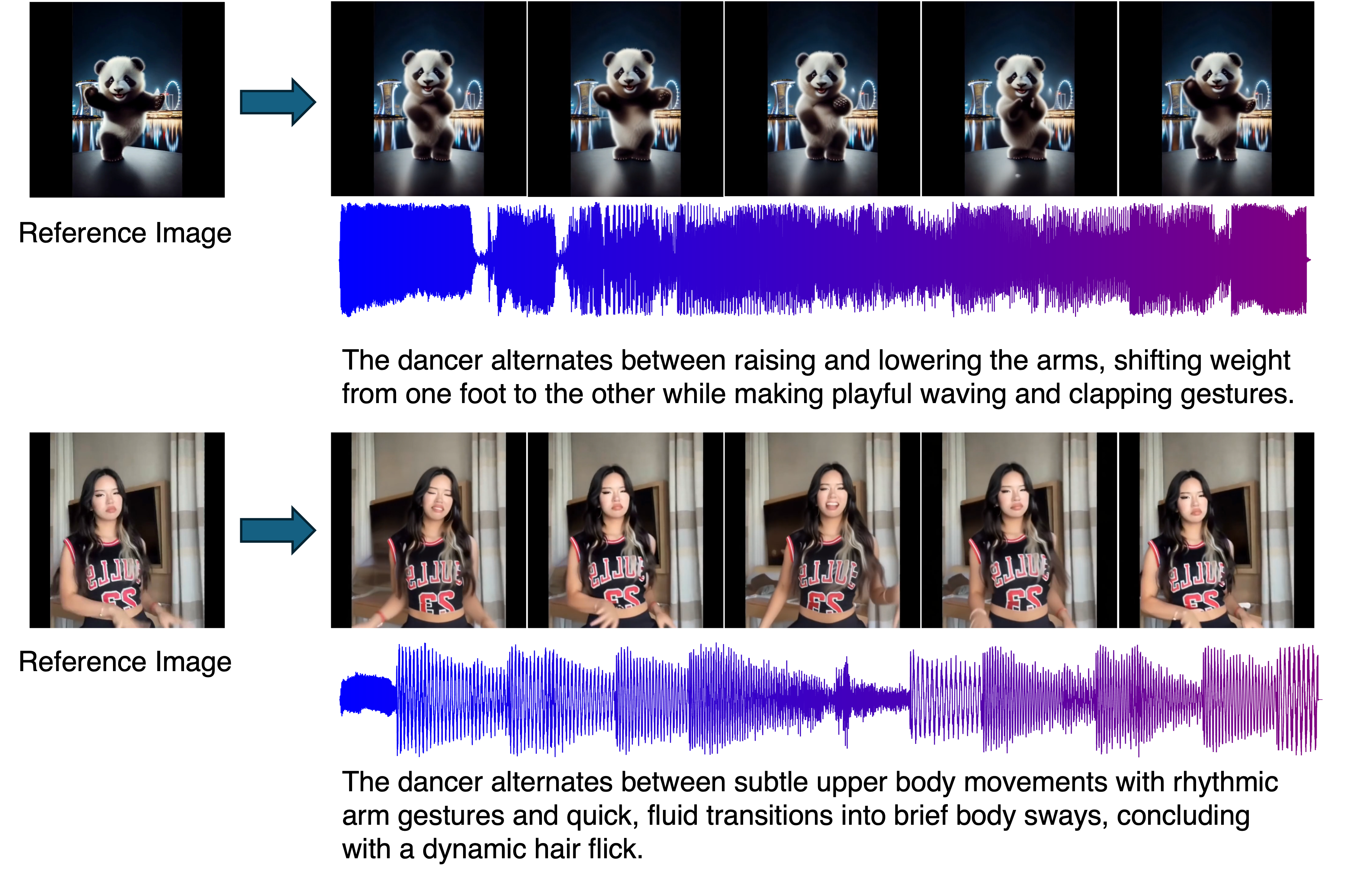}
  \caption{MuseDance generates a dancing video from a reference image, synchronizing movements to the provided music, aligning with the beats, and visually interpreting the guidance of a text prompt for a seamless, music-driven animation.}
  \label{fig:teaser}
\end{figure*}

Artificial Intelligence has made significant strides across various domains recently \cite{brown2020language, touvron2023llama, thoppilan2022lamda, podell2023sdxl, rombach2022high,dong2024musechat,liu2024tackling,dong2024mamba,dong2023cp,dong2024mapping,zhao2024hedge}. 
In particular, image animation have advanced through the use of various of guidances, such as motion~\cite{siarohin2019first, siarohin2021motion, zhao2022thin}, text \cite{chen2025livephoto}.
These methods animate static images by applying motion to main characters within them. However, the area of music-guided image animation remains relatively underexplored. This task involves animating characters to dance in sync with music, ensuring the video aligns with the musical theme and follows the beats. For example, fast-paced pop music tracks suit street dance, while calm piano music complements classical waltz performances. Additionally, we aim to enhance customization by incorporating user-provided text. Even with the same appearance and background, users can express different dance motions guided by the music, creating personalized content that significantly enhances user engagement and amplifies the reach of the music.
This approach has huge potential for widespread applications in industries like film, social media, and education, where users frequently create videos choreographed to popular music.

Several challenges in image animation, particularly in dance video generation, remain unresolved. The primary issues are: (1) the scarcity of large-scale datasets containing dance videos with a variety of subjects—not limited to human dancers—synchronized with background music and accompanied by corresponding text captions; (2) the lack of flexible control over the generation process; and (3) limited generalization beyond the training dataset’s scope. Current motion transfer models rely on motion guidance inputs such as pose, video, depth, and skeleton-based sequences to animate reference images. However, these methods often fail to meet user preferences. For instance, finding suitable dance poses for classic pieces by Beethoven or Mozart can be challenging. The availability of such guidance is limited and often requires specialized domain knowledge to create, which most users lack. Additionally, existing dance motion transfer models focus predominantly on human motion, as the guidance is generally designed for human figures, thereby restricting creativity. In contrast, any object—such as animated characters in Disney movies—could potentially ``dance''. To address these limitations, we propose an end-to-end music-guided dance generation model that animates the reference image using only music and text. This approach enables users, regardless of expertise, to create diverse dance videos featuring not only humans but also a wide range of objects.

In this paper, we introduce MuseDance, a flexible, end-to-end multimodal image animation framework that generates dance videos from a static reference image, a music piece, and a descriptive text prompt of the desired motions. As shown in Figure \ref{fig:teaser}, MuseDance allows the figure in the reference image to dance rhythmically to the music with specified movements, and the video can extend seamlessly based on prior frames. Built on a diffusion-based model, MuseDance adopts the architecture and pretrained weights of Stable Diffusion \cite{blattmann2023stable} and incorporates modifications to ReferenceNet \cite{guo2023animatediff} to capture spatiotemporal information, linking music features and text guidance to create coherent, dynamic animations. The model’s training consists of two stages: in the first stage, the model learns single-frame generation by focusing on the reference image’s appearance, aided by DensePose \cite{guler2018densepose}, and disentangles appearance and motion using text prompts. In the second stage, it learns to create video sequences using the music, beat, and motion guidance, with frozen modules from the first stage to preserve the appearance quality. Three new modules—music, beat, and motion—enable semantic alignment of music features with video sequences, temporal alignment with beats, and consistency across frames, resulting in engaging and dynamic animations.

In summary, our contributions are threefold: (1) We introduce a novel music-dance video generation dataset, where each sample includes a dance video, the corresponding background music track, and the textual description of the motion in the video. This dataset enables models to learn motion dynamics from both music and text, which can be used to animate a reference image. (2) We propose MuseDance, a novel end-to-end diffusion-based method that uses music and text as driving dynamics to animate the reference image in a way that aligns with the semantic meaning of the input. The generated content can be easily modified based on the input data. (3) Our model demonstrates robust generalization, flexible and precise control, as well as the ability to generate nuanced, temporally consistent dance videos across a wide range of objects. We establish this as a baseline for the emerging task of music-guided video generation.

\section{Related Work}
\label{sec:related}

{\bf Video Generation Diffusion Model.} Video generation is a very important task in AIGC. Methods like variational RNNs \cite{babaeizadeh2017stochastic, castrejon2019improved, denton2018stochastic, lee2018stochastic} and GANs \cite{skorokhodov2022stylegan, saito2017temporal, villegas2017decomposing, vondrick2016generating, yu2022generating, li2018video, pan2017create} have been explored to tackle this problem. However, most of those works are limited to low-resolution, the lack of large scale high-quality datasets or loose control ability. Diffusion models are proposed to solve this problem. \cite{blattmann2023align} introduces temporal dimension to the latent diffusion image generation model. Make-A-Video \cite{singer2022make} enhances DALL·E2 \cite{ramesh2022hierarchical}, a text-to-image model, by using joint text-image priors and super-resolution strategies to produce high-quality videos. Stable Video Diffusion \cite{blattmann2023stable} presents a large-scale text-to-video foundation model, which also supports various downstream tasks like image-to-video generation, camera motion adaptability and multi-view objects synthesis. In addition to open source models, closed-source video generative models, in particular
GEN-2 \cite{esser2023structure}, PikaLabs \cite{pika}, Sora \cite{videoworldsimulators2024} and Kling \cite{kling} provide state-of-the-art video generation capabilities for general use.

\noindent{\bf Music-guided Dance Generation.} Music-guided dance generation in 3D sequences has been explored in recent works. Bailando \cite{siyao2022bailando} proposes a pose Q-VAE with a motion GPT to predict future pose tokens given music. EDGE \cite{tseng2023edge} presents a physics-constrained transformer-based diffusion to generate more realistic 3D dance sequences. M2C \cite{marchellus2023m2c} introduces a
music code extractor to replace existing music feature processor to enhance music's role in 3D dance motion generation. LM2D \cite{yin2024lm2d} integrates lyrics information to enable the generation of more diverse 3D dance sequences. 3D dance sequence generation models produce only skeleton keypoints, rather than full dance videos, limiting their practical application. Music-guided dance generation in 2D videos is still largely unexplored. \cite{wang2024dance} utilizes a diffusion model to generate optical flow, which is then combined with a reference image for animation. However, their approach operates at low resolution and has not been tested on specific, large-scale datasets, leading to lower diversity in 2D dance video generation. Our work aims to address these limitations.

\noindent{\bf Human Motion Transfer.} Earlier works \cite{bregler2023video, efros2003recognizing, cheung2004markerless, xu2011video, beier2023feature} on human body motion transfer demonstrate lower accuracy and require significant human intervention. Recently, deep learning techniques have enabled more realistic motion transfer with highly automated training pipelines. MoCoGAN \cite{tulyakov2018mocogan} introduces an unsupervised adversarial training method for transferring motion and facial expressions onto target subjects. \cite{albahar2021pose} extends the StyleGAN \cite{karras2020analyzing} generator to learn the warped local features. \cite{chan2019everybody} utilizes a video-to-video synthesis method to generate new motions by giving a 2D video and 2D skeleton sequences. Dreampose \cite{karras2023dreampose} proposes a diffusion model to animate a reference human image using a target pose sequence and fabric textures. \cite{hu2024animate} utilizes a lightweight pose guider to enable controllable continuous character movement across various downstream tasks. 

In human dance transfer domain, DISCO \cite{wang2024disco} generates human dance videos from dance skeleton sequences and a reference human image. Their method generalizes to unseen human references, backgrounds, and poses. MagicAnimate \cite{xu2024magicanimate} combines a video encoder with an appearance encoder to generate temporally consistent dance videos from a reference image. MagicPose \cite{chang2023magicpose} incorporates facial keypoints with body skeletons as guidance to generate realistic human dance videos. However, these approaches still require pose guidance to generate dance videos, and such pose sequences are not always available, limiting the generalization capability of these models. Our approach aims to address this limitation.

\section{Method}
\label{sec:method}
\begin{figure*}[!h]
  \centering
  \includegraphics[width=.98\textwidth]{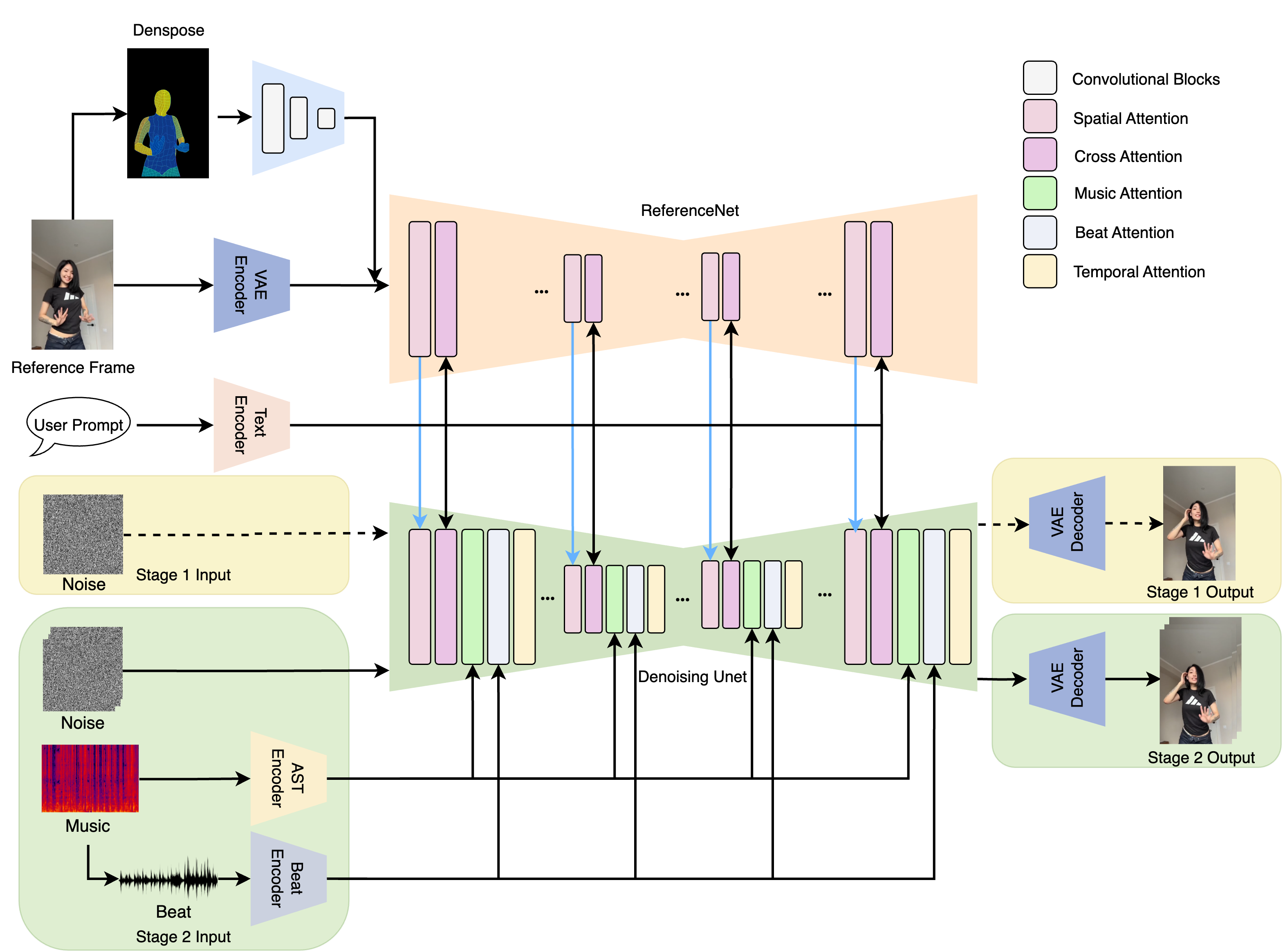}
  \caption{In the first training stage, we train the model to capture spatial information by generating individual frames, with reference and target frames randomly sampled from a short time window. DensePose is used to help the model focus on the object, while text prompts assist in understanding motion. In the second training stage, we freeze the spatial attention blocks to preserve the model’s frame generation ability and introduce music, beat, and motion modules to incorporate music dynamics, align with the beat, and improve frame-to-frame consistency.}
  \label{fig:stage2}
\end{figure*}
We propose a two-step training framework to animate images in dancing based on music and text input. In the first step, the model is trained on individual frames from the target video to learn visual features and acquire prior knowledge. In the second step, we introduce music and text as triggers to generate animated frames that align with these inputs. The process is illustrated below.

\subsection{Preliminaries}
\noindent{\bf Latent Diffusion Models} denote a class of diffusion models that operate within the encoded latent space produced by an autoencoder, represented as $\mathcal{D}(\mathcal{E}(\cdot))$. One of the most widely used models in this category is Stable Diffusion \cite{rombach2022high}, which incorporates a VQ-VAE and a time-conditioned U-Net architecture. Additionally, Stable Diffusion utilizes a text encoder from the CLIP \cite{radford2021learning} model to encode text prompts into embeddings. Given an image $I \in \mathbb{R}^{H_I \times W_I \times 3}$ and its corresponding text embedding $c_{\text{text}} \in \mathbb{R}^{D_c}$, we obtain the latent representation $z_0=\mathcal{E}(I) \in \mathbb{R}^{H_z \times W_z \times D_z}$ and apply it to a predefined diffusion process across $T$ timesteps, modeled as a Gaussian process. This process approximates a standard Gaussian distribution, $z_T \sim \mathcal{N}(0, I)$. The training objective in Stable Diffusion is to iteratively denoise $z_T$ back to the original latent representation $z_0$. 

During inference, the original latent $z_t$ is reconstructed using sampling methods, such as denoising diffusion implicit models \cite{song2020denoising}. Then, the latent $z_t$ is decoded by the decoder $\mathcal{D}$ to generate the final, clear image. Latent Diffusion Models can produce high-fidelity images and align the generated images with the text-conditioned prompt.

\noindent{\bf Cross Attention Mechanism} is a key component in the U-Net of latent diffusion models. This mechanism integrates information from the latent representation and the conditioning embedding, enabling latent diffusion models to generate images that semantically align with the given condition. More generally, the conditioning modality can be text, motion flows, audio, etc., providing semantic guidance for the generation process.

\noindent{\bf ControlNet} \cite{zhang2023adding} is a structure designed to control the generation process without modifying the pretrained diffusion model. ControlNet duplicates the original parameters into a trainable copy while freezing the original weights, allowing the model to retain its generalization ability from large-scale image datasets while fine-tuning the trainable copy for the specific task. The two copies of the model are connected through zero convolution layers, which progressively integrate the information flow. ControlNet enables precise control over image generation by conditioning diffusion models on additional inputs such as edges, poses, or segmentation maps, allowing for more targeted and customizable outputs while maintaining high image quality.

\subsection{Appearance Pretraining}

As shown in Figure \ref{fig:stage2}, the goal of the first training stage is to generate motions similar to those in the reference image while preserving the same appearance. Specifically, we randomly select one frame as the input image and another frame within a fixed time window as the target image. Given a video with $N$ frames, denoted as $V = {I_1, \ldots, I_N}$, we randomly sample a frame $I_i$ as the input image, where $i \in (w, N-w)$, with $w$ representing the fixed time window. We then randomly select a frame $I_j$ as the target image, where $j \in (i-w, i+w)$ and $j \neq i$. Using the DensePose \cite{guler2018densepose} mask, we obtain a dense and robust pose signal of the dancer in the reference image, denoted as $D_i$. We employ a mask encoder $F_m$, which consists of a series of convolutional layers to progressively downsample and extract mask features, yielding $m_i = F_m(D_i)$. Similar to ControlNet \cite{zhang2023adding}, we incorporate mask condition features in a residual manner.

\begin{equation}
z_0=\mathcal{E}(I_i) + \text{Conv}(D_i),
\label{eq:condition_add}
\end{equation}

Similar to approaches in \cite{hu2024animate, xu2024hallo}, we employ a ReferenceNet—a U-Net-based Stable Diffusion model with the same layers as our backbone diffusion model—to extract visual information from the reference image, denoted as $R$. Suppose we have features from the denoising U-Net, $x_d \in \mathbb{R}^{H_z \times W_z \times d}$, and features from ReferenceNet, $x_R \in \mathbb{R}^{H_z \times W_z \times d}$. First, we concatenate these two features along the $W$ dimension, then pass this concatenated feature through spatial self-attention layers, extracting the first half as the output. We then apply this output to cross-attention layers to learn the semantic meaning provided by the CLIP text encoder, denoted as $\operatorname{CrossAttn}(z_t, c_{\text{text}})$. This approach enables the model to learn pose and appearance in the selected frames with high-level semantics.

Since ReferenceNet has a structure similar to that of the denoising U-Net, it can generate feature maps that integrate into the denoising U-Net, enhancing the quality of generated frames with detailed visual information of both foreground and background. This approach differs from ControlNet, which emphasizes alignment between the reference frame and the target frame. In our case, however, the reference and target frames are two different images from a short time window, meaning they share only spatially related features rather than alignment. This is why we chose not to use ControlNet.

\subsection{Dynamic Trigger Video Generation}

Figure \ref{fig:stage2} illustrates our second-stage training process. Here, the model learns to generate dance videos based on the reference image, music input, and text guidance. To preserve the visual generation ability from the appearance pretraining stage, we freeze the spatial attention blocks.

To achieve temporal alignment in driving the reference image, we add three new modules to the denoising U-Net: a music understanding module, a beat alignment module, and a motion alignment module.

\noindent{\bf Music Understanding Module}. This module aims to extract musicality information from the music and use it to control frame generation. We use the AST \cite{gong2021ast} model to obtain the music embedding. Given the hidden states from the previous module, $z_t \in \mathbb{R}^{K \times (H_z W_z) \times d}$, where $K$ is the number of generated frames, and the music embedding $c_{\text{music}} \in \mathbb{R}^{L \times d}$, where $L$ is the sequence length of the music embedding, we apply a cross-attention mechanism between the music embeddings and frames to facilitate information flow across these two modalities, allowing the music dynamics to control frame generation. To further improve temporal alignment, we reshape the hidden states into $z_t \in \mathbb{R}^{(H_z W_z) \times K \times d}$ and compute self-attention along the temporal dimension.

\noindent{\bf Beat Alignment Module}. We observe that, in most music dance videos, the music beat serves as a strong signal, often marking the start, stop, or change in dance style. To capture this pattern, we incorporate beat information into the denoising U-Net. We use Librosa to identify beat locations in the music, converting them into a one-hot encoded format. This produces a binary vector $b_{\text{binary}} \in \mathbb{R}^K$, where frames with a beat are assigned a value of 1, and others a value of 0. We align the beat information with video sequences, inspired by token processing in NLP tasks, and apply a lookup embedding layer to transform the discrete embedding into a continuous dense embedding $b_{\text{dense}} \in \mathbb{R}^{K \times d}$. We then apply the cross-attention mechanism to help the hidden states learn the beat information. Similar to the music understanding module, we reshape the hidden states and apply temporal attention layers to ensure temporal continuity.

\noindent{\bf Motion Alignment Module}. In video generation, maintaining content continuity across frames is crucial, especially for generating coherent dance motions. In addition to the temporal attention layers in the music understanding and beat alignment modules, we employ a motion alignment module to capture temporal dependencies across frames. Inspired by \cite{xu2024hallo, guo2023animatediff}, we use several previously generated frames as guidance, concatenating them with the current hidden states and performing self-attention across the temporal sequence dimension. Specifically, we form a concatenated hidden state $z_{\text{motion}} = \text{concat}(z^\prime_t, z_t)$, where $z^\prime_t \in \mathbb{R}^{(H_z W_z) \times M \times d}$ represents the hidden states from the previous $M$ generated frames. By applying self-attention on $z_{\text{motion}}$ across the temporal axis, we select the last $K$ hidden states as the current generated frames.
\section{Experimental Results}
\label{sec:experiments}
\begin{figure*}[!h]
  \centering
  \includegraphics[width=0.8\textwidth]{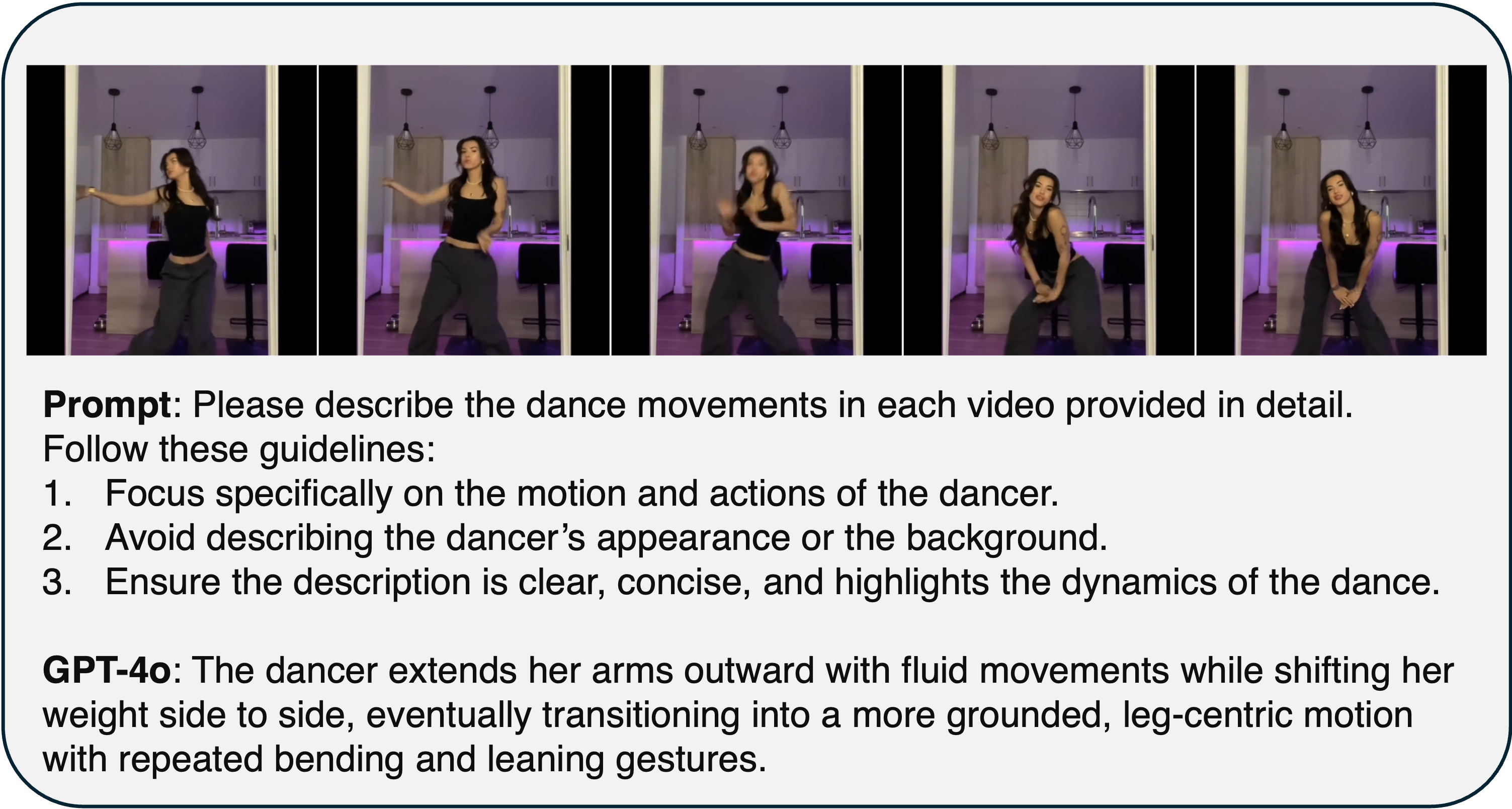}
  \caption{An example of textual data generation, we provide a series of frames and a detailed prompt to instruct GPT-4o to generate motion captions.}
  \label{fig:prompt}
\end{figure*}
\subsection{Music Dance Dataset}
In this paper, we introduce the first music-dance video dataset. The raw videos are collected from YouTube, totaling 304 videos. These videos feature a diverse range of dance genres, from popular styles found on short video platforms to traditional Chinese dance. The videos also provide diversity in dancer appearance and background settings. Given that dancers often perform to the same music, some music tracks overlap across videos. Human-object videos are primarily from TikTok dance collections or dance challenge series, while non-human object videos consist of synthetic animations of animals dancing. 
We split each video into multiple clips. All these
 sub-videos are in a vertical, object-centered format, paired with background music, and vary in length from 10 to 30 seconds. We manually edit each video to remove irrelevant segments, such as intros, outros, and conversational parts, retaining only the music-driven dance sections. To ensure consistency, we include only videos with a single dancer, with plans to expand the dataset to include multiple-dancer videos in the future. Following these preprocessing steps, our dataset comprises 2,904 videos, each paired with background music and lasting approximately 4 seconds. The dataset includes a total of 454 unique music tracks. All dance videos undergo manual review to ensure that any harmful or inappropriate content is excluded.

We also include a text description of motion for each pair of music and video. Figure \ref{fig:prompt} illustrates our process for generating these descriptions. We leverage OpenAI’s GPT-4o API to generate video captions, sampling each video every 10 frames and combining these samples with a text prompt for GPT-4o. To better disentangle motion from appearance, we instruct GPT-4o to ignore the dancer’s appearance and background, focusing solely on motions and actions. Under this setup, each sample in our dataset includes a triplet: a short dancing video, background music, and a motion description text.

In Table \ref{tab:dataset_comparison}, we compare existing music-dance datasets with our own. The Everybody Dance Now \cite{chan2019everybody} and TikTok \cite{jafarian2021learning} datasets contain only the video modality and have significantly smaller dataset sizes. The AIST++ \cite{li2021ai} dataset, a subset of the AIST  \cite{tsuchida2019aist} dataset, includes both video and music modalities. However, it is limited to 10 dance genres and 60 music tracks, resulting in less diversity in music and videos compared to our dataset. Additionally, its video backgrounds are clear and simple, which may lack the richness and realism needed for creating diverse dancing videos with complex backgrounds.

\begin{table}[!t]
  \centering
  \resizebox{1\linewidth}{!}{
  \begin{tabular}{@{}llll@{}}
    \toprule
    Dataset & Videos & Music & Text \\
    \midrule
    Everybody Dance Now \cite{chan2019everybody} & 105 Videos & \xmark & \xmark \\
    TikTok \cite{jafarian2021learning} & 350 Videos & \xmark & \xmark \\
    AIST++\tablefootnote{For a fair comparison, we only consider videos filmed from a front-facing camera perspective with a single dancer.} \cite{li2021ai} & 1408 Videos & 60 Songs & \xmark \\
    MuseDance (ours) & 2904 Videos & 454 Songs & 2904 Captions \\
    \bottomrule
  \end{tabular}
  }
  \caption{Current Music Dance Dataset Comparison.}
  \label{tab:dataset_comparison}
\end{table}

\subsection{Implement Details}
In our experiments, both training and inference processes are conducted on a computational platform with 32 NVIDIA A100 GPUs, each has 80 GB memory. The training framework consists of two stages, each comprising 30,000 steps. The batch size is set to 12 in the first stage and 2 in the second stage for each GPU, with video dimensions maintained at 640 × 640 pixels. During the second stage of training, each instance generates a 4-second video at a frame rate of 12 fps. The music has a sample rate of 16,000 Hz and is in mono. To ensure consistency in the generated content, the hidden states of the last two generated frames are utilized within the motion module. A learning rate of 1e-5 is applied across both training stages, and the Adam optimizer is employed for parameter updates. The ReferenceNet and Denoising U-Net are initialized based on \texttt{stable-diffusion-v1-5}, while the motion module is initialized with weights derived from Animatediff \cite{guo2023animatediff}. To enhance video generation quality, a dropout rate of 0.05 is applied. Additionally, we use DDIM to sample the generated frames.
\begin{figure*}[!h]
  \centering
  \includegraphics[width=1\textwidth]{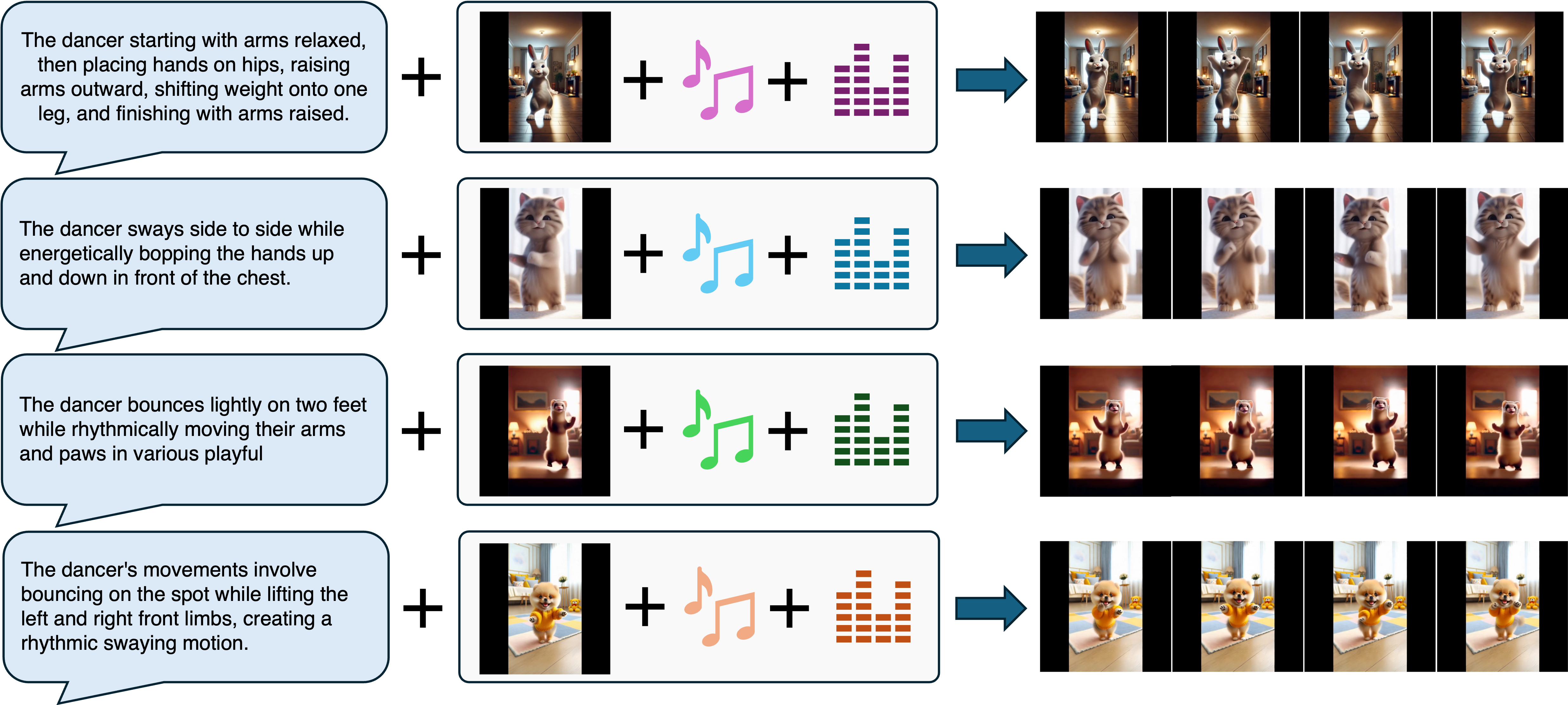}
  \caption{Music driven dancing video generation on non-human objects.}
  \label{fig:non_human}
\end{figure*}

\subsection{Quantitative Results}
Similar to the approach in \cite{wang2024disco}, we randomly select 10 videos as the test split, including various figures, such as human and non-human ones. We evaluate the quality of our generated dancing videos using several metrics. For single-frame quality, we employ SSIM \cite{wang2004image}, LPIPS \cite{zhang2018unreasonable}, and PSNR \cite{hore2010image}. To assess overall video quality, we use Fréchet Video Distance (FVD) \cite{unterthiner2018towards}. 

Due to the lack of open-source code and data for the music-driven video generation task, we don't have baseline models for direct comparison. To evaluate our results, we compare our results from two different perspectives: (1) We select a pair of models for comparison. Firstly, we use EDGE \cite{tseng2023edge}, generates 3D skeleton sequences based on music input. And we project those sequences into 2D space. Using these 2D pose sequences as guidance, we animate the reference image according to the approach in DISCO \cite{wang2024disco}. (2) We leverage MM-Diffusion \cite{ruan2023mm}, a joint audio-video generation model, using the music in the MuseDance dataset as the condition to generate dancing videos. We reimplement those models on our manually collected dataset and Table \ref{tab:comparison} shows the comparison results.

\begin{table}[h]
\centering
\resizebox{\columnwidth}{!}{%
\begin{tabular}{lcccc}
\hline
\multirow{2}{*}{Method} & \multicolumn{3}{c}{Image} & \multicolumn{1}{c}{Video} \\
\cline{2-4} \cline{5-5}
& PSNR $\uparrow$ & SSIM $\uparrow$ & LPIPS $\downarrow$ & FVD $\downarrow$ \\
\hline
EDGE + DISCO & 27.68 & 0.602 & \textbf{0.266} & 396.95 \\
MM-Diffusion & 28.81 & 0.469 & 0.297 & 586.87 \\
MuseDance (ours) & \textbf{29.59} & \textbf{0.680} & 0.276 & \textbf{311.04} \\
\hline
\end{tabular}%
}
\caption{Comparison of models' performance across image and video metrics on MuseDance dataset.}
\label{tab:comparison}
\end{table}

\subsection{Ablation Studies}
To illustrate the effectiveness of each module in the second training stage, we conduct ablation studies by removing the music module, beat module, or motion module to examine their influence on the generated content and consistency across frames. For a fair comparison, we freeze the modules trained at the first stage. As shown in Table \ref{tab:ablation}, we observe that adding the music, motion, and beat modules effectively enhances generation quality. Notably, the motion module significantly improves temporal consistency.
\begin{table}[h]
\centering
\resizebox{\columnwidth}{!}{%
\begin{tabular}{ccccccc}
\hline
\multicolumn{3}{c}{Module} & \multicolumn{4}{c}{Metrics} \\
\cline{1-3} \cline{4-7}
Music & Motion & Beat & PSNR $\uparrow$ & SSIM $\uparrow$ & LPIPS $\downarrow$ & FVD $\downarrow$ \\
\hline
\xmark & \xmark & \xmark & 24.49 & 0.611 & 0.291 & 738.69 \\
\cmark & \xmark & \xmark & 24.63 & 0.619 & 0.283 & 612.34 \\
\cmark & \cmark & \xmark & 28.88 & 0.674 & 0.277 & 400.55 \\
\cmark & \cmark & \cmark & \textbf{29.59} & \textbf{0.680} & \textbf{0.276} & \textbf{311.04} \\
\hline
\end{tabular}%
}
\caption{Ablation study results showing the impact of removing different modules in the second training stage.}
\label{tab:ablation}
\end{table}

\subsection{Qualitative Results}
{\bf Non-human Object Generation.} Unlike existing works, our model has the capability to generate dancing videos of non-human objects. As shown in Figure \ref{fig:non_human}, our model produces realistic dancing motions for non-human objects based on the music input and tempo. We observe that regardless of whether the text description provides detailed motion instructions or just generalized guidance, the model still performs well. This demonstrates the strong language understanding ability of our model.

\begin{figure}[!h]
  \centering
   \includegraphics[width=0.9\linewidth]{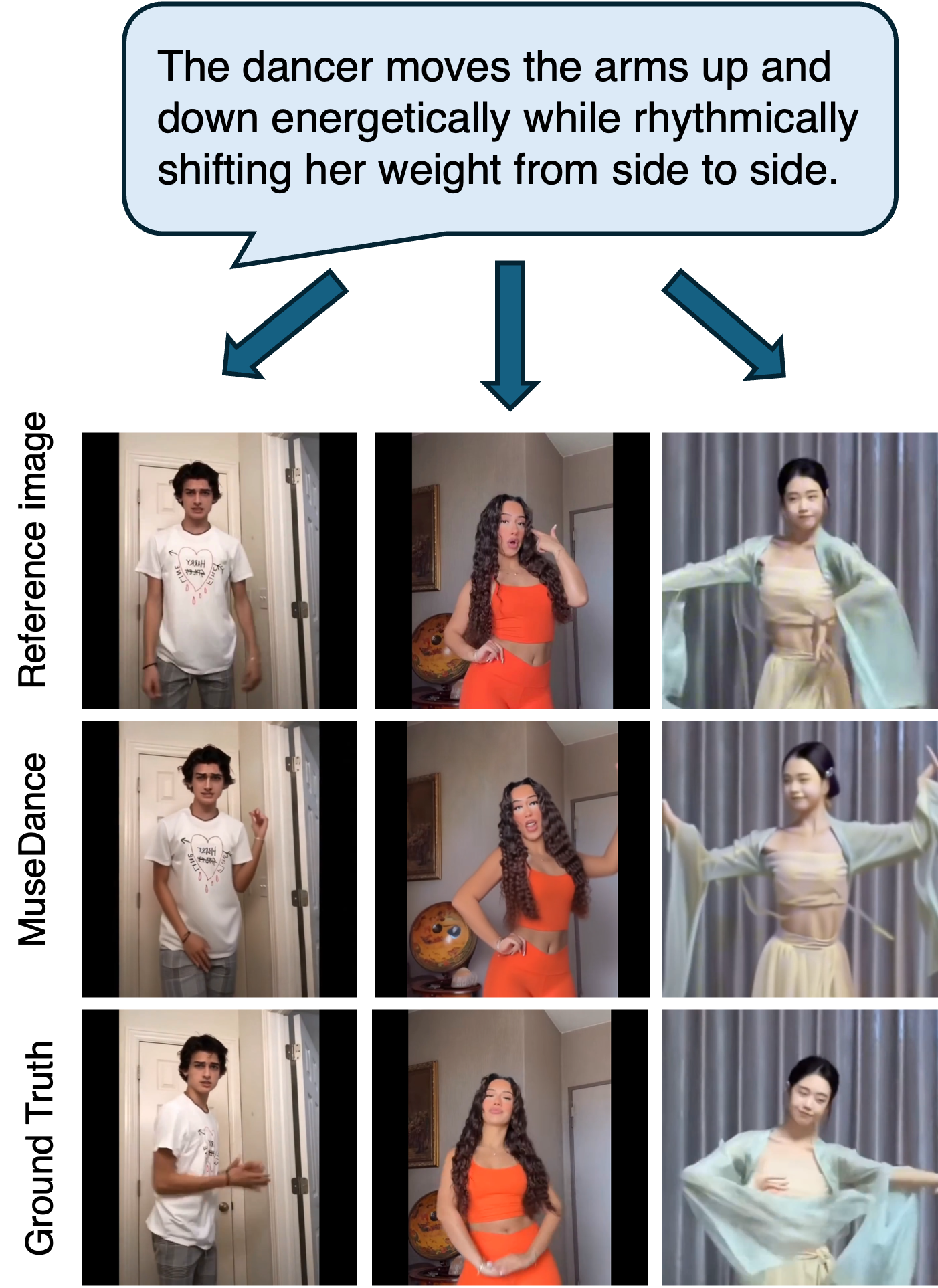}
   \caption{Dance video generations with the same text prompt but different reference images and music inputs. Frames are shown at the same time points from both the generated videos and the ground truth.}
   \label{fig:same_text}
\end{figure}

\noindent{\bf Text Semantic Preservation.} We evaluate our model’s ability to preserve semantic consistency by controlling the input text prompt. In Figure \ref{fig:same_text}, we control the text guidance and use different music dynamics to drive various reference images. The results demonstrate that our model can accurately interpret the text guidance while maintaining the flexibility to incorporate the music input and generate coherent dance videos.

\section{Potential Applications}
\label{sec:potential_app}
{\bf Dynamic Content Creation in Social Media and Interactive Marketing.} MuseDance’s ability to animate a wide range of objects, not limited to human figures, in sync with musical and textual guidance opens up new possibilities in social media content creation and interactive marketing. Users can create engaging videos featuring personalized or branded objects that ``dance" to trending music tracks, significantly enhancing engagement and shareability. This unique flexibility extends beyond traditional motion transfer applications and can cater to users’ need for quick, dynamic content on platforms like TikTok, YouTube and Instagram.

\noindent{\bf Interactive Virtual Environments and Gaming.} MuseDance can introduce lifelike animation to various in-game assets or virtual environments, where objects or characters move rhythmically with background music, creating an immersive atmosphere. For instance, in rhythm-based games or virtual concerts, MuseDance can animate avatars, objects, or even abstract art elements in response to the music track, amplifying user immersion and engagement. Unlike existing methods that require explicit pose guidance, MuseDance’s reliance on music and text input makes it highly adaptable, enabling the seamless generation of synchronized movements for any character or object within the virtual scene.

\noindent{\bf Enhanced Music and Dance Education Tools.} MuseDance’s capacity to synchronize animations with diverse types of music and rhythm patterns could be harnessed in educational platforms focused on music and dance. By animating reference images based on music, the model can help students visualize how various music styles—ranging from classical to jazz or pop—affect dance movements. This visual and interactive aid can serve as a practical tool for students to study the influence of rhythm and melody on motion, supporting not only traditional dance education but also aiding musicians in understanding dance as a form of physical expression of music.
\section{Limitations}
\label{sec:limitations}
Although our framework allows for the generation of videos of unlimited length, extending videos for too long can reduce frame consistency and compromise the realism of the content. As the sequence progresses, minor inconsistencies may accumulate, leading to flickering or unrealistic transitions. Additionally, while our dataset includes detailed text descriptions that capture body motions sequentially, these descriptions lack explicit temporal information. This limitation affects the model’s ability to disentangle appearance and motion effectively, which may impact the temporal consistency of the generated video. In future work, we plan to expand our dataset with richer text annotations that include explicit timeline information, providing clearer temporal guidance. This enhancement could improve the model’s ability to separate appearance and motion, resulting in more coherent and consistent video sequences.
\section{Conclusions}
\label{sec:conclusion}
In this work, we explore the potential of using music dynamics and text guidance to animate static images, creating an end-to-end framework for flexible dance video generation. We developed the first music-driven dance video generation dataset from publicly available YouTube videos and proposed a diffusion-based model capable of integrating visual, auditory, and textual cues. Our model not only ``sees" the reference image but also ``listens" to music, senses tempo and beat, and interprets text instructions to produce realistic, synchronized dance videos. Extensive experiments demonstrate the strong performance of our model on this novel multimodal task. We have also outlined potential applications and acknowledged current limitations, which we aim to address in future work. We hope this study sparks further exploration and innovation in this exciting area.
{
\small
\bibliographystyle{ieeenat_fullname}
\bibliography{references}
}


\end{document}